\documentclass[10pt]{article} 

\usepackage[accepted]{rlj} 

\usepackage{amssymb}
\usepackage{mathtools}
\usepackage{mathrsfs}
\usepackage{graphicx}
\usepackage{subcaption}
\usepackage[space]{grffile}
\usepackage{url}
\usepackage{booktabs}
\usepackage{multirow}
\usepackage{algorithm}
\usepackage{algorithmic}
\usepackage{amsthm}
\usepackage{thmtools}

\definecolor{dred}{RGB}{153,80,43}
\hypersetup{linkcolor=dred}

\usepackage[capitalize,noabbrev,nameinlink]{cleveref}
\crefname{appendix}{Appendix}{Appendices}
\creflabelformat{equation}{#2#1#3}

\title{RENEW: Towards Learning World Models and \\ Repairing Model Exploitation from Preferences}

\setrunningtitle{RENEW: Towards Learning World Models and Repairing Model Exploitation from Preferences}

\author{
Logan Bhamidipaty\textsuperscript{1}, 
Mykel Kochenderfer\textsuperscript{2}, 
Subramanian Ramamoorthy\textsuperscript{1}}

\emails{l.m.bhamidipaty@sms.ed.ac.uk}

\affiliations{
$^{1}$\textbf{University of Edinburgh}\\
$^{2}$\textbf{Stanford University}\\
}

\keywords{world models, model-based RL, model exploitation, offline RL, RLHF} 

\begin{document}

\maketitle  

\begin{abstract}
World models are widely used in offline reinforcement learning (RL) to improve sample efficiency and generate experience beyond a fixed dataset. However, they are vulnerable to model exploitation where data coverage is thin. Prior work addresses this either by collecting more expert demonstrations, which is often expensive, unsafe, or unavailable, or by conservative algorithms that avoid uncertain regions, which limits generalization. We propose instead to repair exploitation directly using human preferences over imagined rollouts, leveraging the strong intuitive physics that allows humans to easily spot egregious dynamics hallucinations. We formalize this as Dynamics Learning from Human Feedback (DLHF), a Bradley-Terry preference loss over trajectory log-likelihoods under a learned dynamics model. Unfortunately, naive DLHF is sample inefficient, so we introduce RENEW, which uses epistemic uncertainty to focus finetuning where the model is most exploitable. We evaluate on several Jumanji and classic control environments and find that while naive DLHF requires an outsize preference budget, RENEW makes the framework practical by improving sample efficiency, limiting catastrophic forgetting, and reducing exploitation in pretrained world models. Taken together, our results provide initial evidence that preferences can supervise world model dynamics directly, offering a new approach to addressing exploitation in offline model-based RL.\footnote{Code: \url{https://github.com/FlyingWorkshop/RENEW}}
\end{abstract}

\section{Introduction}

Model-based reinforcement learning offers a compelling path toward sample-efficient decision-making: by learning a dynamics model of the environment, an agent can plan and optimize behavior without costly real-world interaction \citep{dyna, deisenroth2011pilco, chua2018deep}. In the offline setting, where the agent must learn entirely from a fixed dataset, learned world models are especially attractive because they allow the agent to generate synthetic experience beyond what the dataset contains, effectively expanding the coverage of limited data \citep{janner2019trust, yu2020mopo}. However, this same capability creates a fundamental vulnerability. When a policy optimizes against a world model, it may discover and exploit inaccurate transitions to achieve artificially high predicted returns. This failure mode, known as model exploitation \citep{ha2018world, janner2019trust, bhamidipaty2026imperfect}, ranges from agents that teleport through walls in a simulated maze to deployed systems that act on incorrect dynamics with serious consequences, as when a robotaxi failed to model a pedestrian trapped beneath its chassis and continued driving while dragging them along underneath \citep{koopman2024anatomyrobotaxicrashlessons}.

The typical response to model exploitation is pessimism: penalize the agent for visiting states where the model is uncertain \citep{yu2020mopo, kidambi2020morel}, or constrain the policy to remain close to the behavioral distribution \citep{fujimoto2019off, kumar2019stabilizing}. While effective in some regimes, these approaches trade off exploration and generalization for robustness, limiting the very capabilities that make model-based methods appealing. A pragmatic alternative is to simply \emph{repair the model} by collecting more data in the regions where it is wrong \citep{dagger}. Yet in many practical domains, additional demonstrations are expensive, dangerous, or unavailable entirely. Expert teleoperation in robotics, near-failure driving data for autonomous vehicles, and rare economic events all represent settings where the training distribution cannot easily be expanded \citep{zolna2020offline, yu2022leverage}. This raises a question: \textit{can we correct a world model's dynamics without collecting more demonstrations?}

We observe that while demonstrating correct behavior requires domain expertise, identifying incorrect dynamics is often trivial. Humans possess strong intuitive models of physics, causality, and object permanence \citep{spelke1992origins, baillargeon2004infants}, and even large-scale generative world models still produce outputs that violate these intuitions: objects pass through solid surfaces, entities appear or vanish between frames, and dynamics contradict basic mechanical expectations \citep{motamed2026generative}. We do not aim to supervise nuanced physical laws such as fluid dynamics or thermodynamics; rather, we target the coarse, obvious violations that arise from model exploitation and that humans recognize immediately and effortlessly \citep{riochet2021intphys}. This motivates a different form of supervision: rather than asking an expert to \emph{demonstrate} correct behavior, we can ask a non-expert to \emph{compare} two trajectory segments and indicate which is more physically plausible. We formalize this idea as Dynamics Learning from Human Feedback (DLHF), replacing the reward model in the Bradley-Terry preference framework \citep{bradley1952rank, christiano2017deep, ouyang2022training} with the trajectory log-likelihood under a dynamics model, so that preference labels supervise transitions directly without requiring an expert policy or environment interaction.

However, learning dynamics from preferences is a harder problem than learning rewards from preferences. In typical RLHF, the preference signal trains a scalar reward function. In DLHF, the same binary comparison must supervise a high-dimensional structured transition model that maps states and actions to full next-state distributions. Each preference label therefore carries less information per parameter, and naively collecting preferences over uniformly sampled rollouts wastes much of the budget on transitions the model already predicts well. To address this, we introduce RENEW (\textbf{R}epairing \textbf{E}xploitatio\textbf{N} with \textbf{E}licited \textbf{W}orld-model preferences), which uses epistemic uncertainty to direct preference queries toward transitions where the model is most uncertain and therefore most vulnerable to exploitation.

We show three results in several Jumanji \citep{bonnet2024jumanji} and classic control \citep{gymnax2022github} environments: (1) binary preferences over trajectory rollouts are sufficient to learn simple world model dynamics from scratch, with no demonstrations, reward labels, or environment interaction; (2) active preference querying via epistemic uncertainty improves sample efficiency over uniform querying and avoids the catastrophic forgetting that naive DLHF exhibits; and (3) under a fixed preference budget, RENEW reduces prediction error and epistemic uncertainty in pretrained world models more effectively than naive finetuning, targeting the conditions that give rise to model exploitation.

\begin{figure}[t]
    \begin{center}
        \includegraphics[]{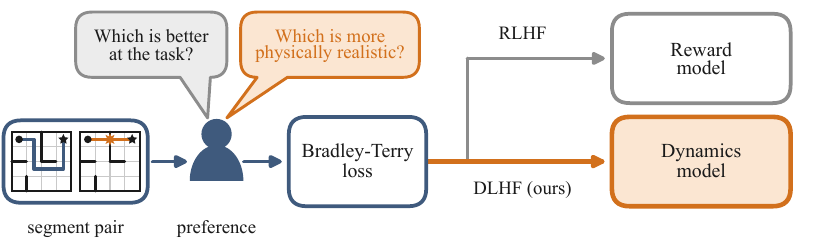}
    \end{center}
    \caption{RLHF and DLHF share the same preference machinery but supervise different components. A human compares a segment pair, and the resulting binary preference is optimized under a Bradley-Terry loss. \textbf{RLHF:} the preference trains a reward model, so labels encode \emph{task success}. \textbf{DLHF (ours):} the trajectory log-likelihood $\ell_\theta$ replaces the reward model (\cref{eq:dlhf}), so the same labels supervise the dynamics model directly and encode \emph{physical realism}. In the segment pair shown, both rollouts share a start ($\bullet$) and goal ($\star$), and the orange rollout reaches the goal in 3 steps rather than 7 by clipping through a wall. Asked which is better at the task, an annotator may prefer the faster trace, reinforcing model exploitation; asked which is more physically realistic, the same annotator may prefer the plausible trace.}
    \label{fig:dlhf}
\end{figure}

\section{Methodology}

We present our approach in two parts.\footnote{For brevity, we defer the preliminaries to \cref{sec:prelims}.} First, we formalize the problem of learning transition dynamics from human preferences, which we call Dynamics Learning from Human Feedback (DLHF). Second, we introduce RENEW (\textbf{R}epairing \textbf{E}xploitatio\textbf{N} with \textbf{E}licited \textbf{W}orld-model preferences), an algorithm that instantiates DLHF to repair model exploitation in offline world models via active preference querying.

\subsection{Dynamics Learning from Human Feedback}

In typical RLHF, human preferences over trajectory segments are used to learn a reward model $R_\phi$ (\cref{sec:rlhf}). We observe that the same preference framework can instead supervise the learning of transition dynamics $\hat{\mathcal{T}}_\theta$ directly. This reframing is motivated by a practical observation: in many domains, the expert demonstrations needed to train world models are expensive, dangerous, or unavailable, while humans possess strong intuitive priors over physical dynamics. Decades of developmental research have shown that even young infants expect objects to obey principles of solidity, continuity, and persistence \citep{spelke1992origins, baillargeon2004infants}, and adults can readily distinguish plausible from implausible state transitions without needing to specify a reward function.\footnote{\citet{riochet2021intphys} provide an accessible summary of the developmental timeline of intuitive physics in infants.} Preferences over imagined trajectories thus offer a cheaper supervision signal for dynamics learning that requires no expert policy and no environment interaction.

It is important that preferences encode \textit{realism} rather than \textit{task success}. To see why, consider a maze-solving agent whose world model allows it to teleport through walls. A human providing reward preferences may prefer the teleporting trajectory, since it reaches the goal faster. A human providing dynamics preferences would reject it, since walls are solid. Reward preferences can therefore reinforce model exploitation rather than correct it. DLHF avoids this failure mode by construction: the preference signal is grounded in physical plausibility, not task performance (see \cref{fig:dlhf}).

\paragraph{Problem formulation.} We consider the reward-free setting introduced in \cref{sec:offline}. Given a dynamics model $\hat{\mathcal{T}}_\theta$, we define the log-likelihood of a trajectory segment $\sigma = (s_0, a_0, s_1, \ldots, s_H)$ under the model as:
\begin{equation}
    \ell_\theta(\sigma) = \sum_{t=0}^{H-1} \log \hat{\mathcal{T}}_\theta(s_{t+1} \mid s_t, a_t)
\end{equation}
The central observation is that $\ell_\theta$ can replace the reward model $R_\phi$ in the Bradley-Terry preference framework (\cref{eq:bt}). Given a pair of segments $(\sigma^0, \sigma^1)$, we model the probability that $\sigma^0$ is preferred as:
\begin{equation}
\label{eq:dlhf}
    P(\sigma^0 \succ \sigma^1) = \mathrm{logistic}(\ell_\theta(\sigma^0) - \ell_\theta(\sigma^1))
\end{equation}
and optimize $\theta$ by minimizing the cross-entropy loss from \cref{eq:rlhf_loss}, substituting $\ell_\theta$ for $R_\phi$:
\begin{equation} \label{eq:dlhf_loss}
    \mathcal{L}_{\text{DLHF}}(\theta) = -\sum_{(\sigma^0, \sigma^1, y) \in \mathcal{D}_\succ} \left[ (1 - y) \log P(\sigma^0 \succ \sigma^1) + y \log P(\sigma^1 \succ \sigma^0) \right]
\end{equation}
The dynamics model itself serves as the preference function: a trajectory is preferred if it is more likely under the learned dynamics. This formulation naturally accommodates latent dynamics models; when $\hat{\mathcal{T}}_\theta$ operates over learned representations $z$ via an encoder $e_\psi$ and latent dynamics $\hat{\mathcal{T}}_\theta(z' \mid z, a)$ (\cref{sec:mbrl}), we replace $\ell_\theta$ with the corresponding latent log-likelihood.

\subsection{RENEW}

\begin{algorithm}[t]
\caption{RENEW}
\label{alg:renew}
\begin{algorithmic}[1] 
\REQUIRE Pretrained world model $\hat{T}_\theta$, preference budget $N$, number of iterations $I$
\FOR{$i = 1, \dots, I$}
    \STATE Compute epistemic uncertainty $u$ under current $\hat{T}_\theta$
    \STATE Sample segment pairs $(\sigma^0, \sigma^1)$ from $\hat{T}_\theta$ proportional to $u$
    \STATE Elicit $N/I$ preferences $\mathcal{D}_{\succ} = \{(\sigma^0_j, \sigma^1_j, y_j)\}_{j=1}^{N/I}$
\STATE Update $\theta \leftarrow \arg\min_\theta \mathcal{L}_{\text{DLHF}}(\theta; \mathcal{D}_{\succ})$ \hfill (\cref{eq:dlhf_loss})
\ENDFOR
\RETURN $\hat{T}_\theta$
\end{algorithmic}
\end{algorithm}

DLHF provides a general formulation for dynamics learning from preferences. However, collecting preferences over uniformly sampled rollouts is inefficient: most trajectory pairs will cover regions where the model is already accurate. In the offline setting, model exploitation arises specifically in regions where the model has high epistemic uncertainty due to poor coverage in the training data $\mathcal{D}$. RENEW targets these regions directly.

We assume access only to the pretrained model, not the original training data. This is practically motivated: the pretraining corpus may be proprietary, prohibitively large, or subject to privacy constraints. RENEW finetunes exclusively on preference data.

\paragraph{Epistemic uncertainty.} Let $u(s, a)$ denote an epistemic uncertainty estimator over the dynamics model at state-action pair $(s, a)$. Multiple estimators have been proposed in the MBRL literature, including ensemble disagreement \citep{lakshminarayanan2017simple}, maximum predicted variance \citep{yu2020mopo}, probabilistic ensembles \citep{chua2018deep}, and bootstrapped uncertainty \citep{osband2016deep}. RENEW is agnostic to the choice of $u$. Intuitively, the active querying loop relies on $u$ to identify regions where the dynamics model is underspecified by the training data, so that preference supervision can be directed where it is most needed. We extend $u$ to trajectory segments by defining $u(\sigma) = \sum_{t=0}^{H-1}u(s_t, a_t)$.

\paragraph{Active preference querying.} At each round, we generate candidate trajectory segment pairs by rolling out the current dynamics model from a pool of valid starting states.\footnote{In practice, we use states from the offline pretraining dataset as the starting pool. However, RENEW does not use the offline data as a training signal during finetuning, only as a source of valid initial conditions from which to generate rollouts for preference labeling. Any source of valid states suffices, including states drawn from the environment's initial state distribution $d_0$.} We sample start states with probability proportional to their mean epistemic uncertainty across actions, concentrating the preference budget on transitions where the model is most likely to be exploited by a downstream planning agent. Actions are sampled uniformly. This soft selection maintains diversity across the queried regions while still biasing heavily toward high-uncertainty states. Alternative selection strategies such as top-$k$ or nucleus sampling~\citep{fan2018hierarchical, holtzman2019curious} over the uncertainty distribution can also be used to balance concentration and coverage.

\paragraph{Finetuning.} After curating the preference dataset, we finetune $\theta$ by minimizing $\mathcal{L}_{\text{DLHF}}(\theta)$ (\cref{eq:dlhf_loss}). The process iterates over $I$ rounds: after each round of finetuning, we recompute the epistemic uncertainty under the updated model, so that subsequent rounds direct the preference budget toward the remaining high-uncertainty regions. This adaptive targeting ensures that supervision is not wasted on transitions the model has already corrected in earlier rounds.

We detail practical considerations in \cref{sec:practical-considerations}.

\section{Experimental Evaluation}

We pose three questions for our experimental evaluation.
\begin{enumerate}
    \item \textit{Can binary preferences serve as a training signal for world model dynamics?} Yes, but at high cost: DLHF requires around 1M preference labels to reach error comparable to 1K supervised transitions (\cref{sec:from-scratch}).

    \item \textit{Does active query selection improve sample efficiency over uniform querying?} Yes. Under identical preference budgets, RENEW roughly halves the final error on Sliding Tile $3 \times 3$ and avoids the catastrophic forgetting that naive DLHF exhibits on Sokoban (\cref{sec:sample-efficiency}).

    \item \textit{Can preference-based finetuning reduce transition errors in regions vulnerable to model exploitation?} Yes. With only 1600 labels, RENEW reduces both prediction error and epistemic uncertainty in pretrained world models (\cref{sec:repair}).
\end{enumerate}

\subsection{Learning Discrete World Models from Preferences}
\label{sec:from-scratch}

\begin{figure}[t]
    \begin{center}
        \includegraphics[width=\textwidth]{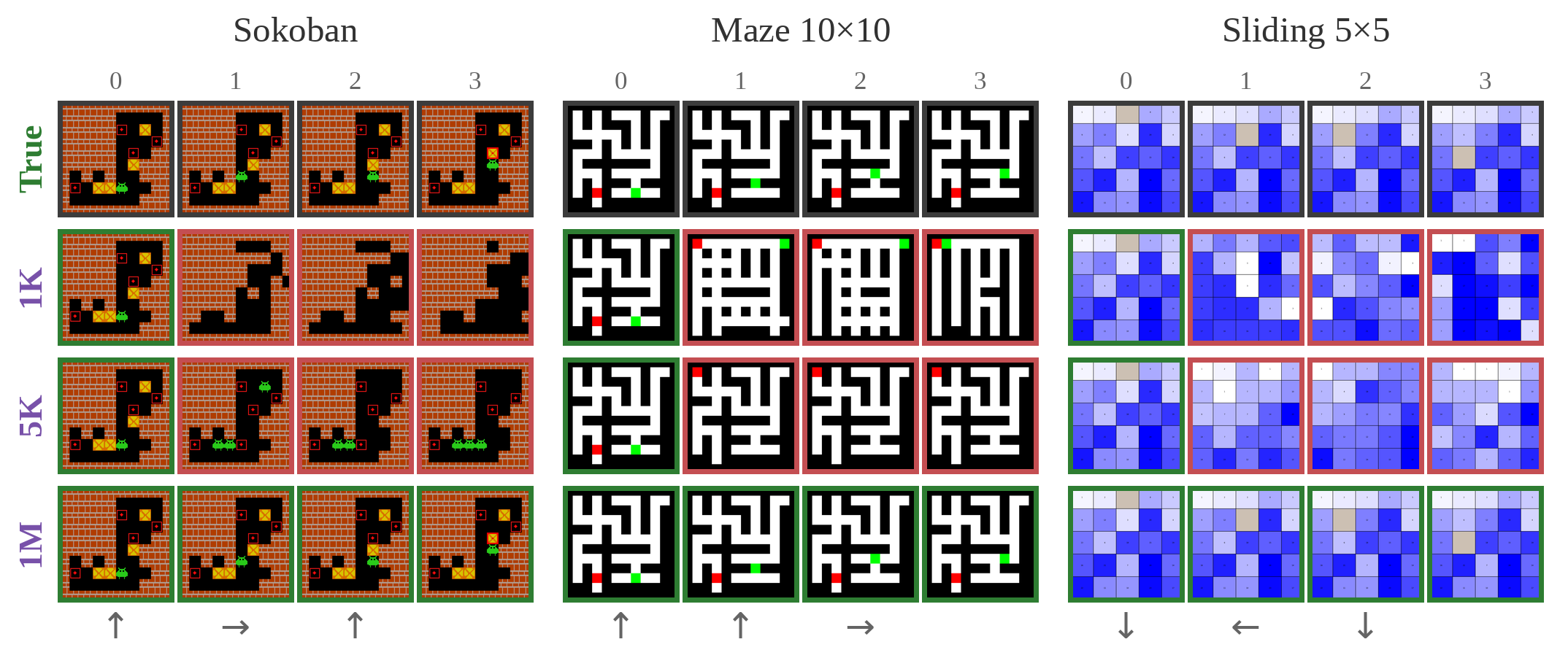}
    \end{center}
    \caption{Imagined rollouts from world models trained from preferences on three Jumanji environments. \textbf{Row 1:} ground-truth dynamics. \textbf{Rows 2 to 4:} model predictions after training on 1K, 5K, and 1M preference labels. \textcolor{green!50!black}{Green} borders indicate correct predictions; \textcolor{red}{red} borders indicate errors.}
    \label{fig:rollouts}
\end{figure}

We first seek a proof of life: \textit{can naive DLHF learn simple world models in an idealized setting?} We use a synthetic oracle that labels preference pairs by $\ell_1$ distance to the ground-truth transition, a large preference budget (up to 1M labels), and small discrete environments from the Jumanji suite \citep{bonnet2024jumanji} (\cref{app:envs}); continuous control results are in \cref{app:classic-control}. We train a convolutional world model with latent dynamics (\cref{app:conv-arch}). If naive DLHF fails under these conditions, the framework has no hope of scaling.

\Cref{fig:rollouts} and \cref{tab:from-scratch} confirm that naive DLHF can learn transition dynamics from scratch across environments. At 1K preferences, the model produces recognizable but incorrect structure; at 1M, predictions match ground truth. However, DLHF requires several orders of magnitude more labels to reach error comparable to 1K supervised transitions. We note that under a synthetic oracle, DLHF is effectively a diluted form of supervised learning.

\begin{table}[h]
    \caption{Final validation $\ell_1$ error for world models trained from scratch. Supervised: 1K random transitions. DLHF: 1M preference labels. Mean $\pm$ 95\% CI over 6 seeds.}
    \label{tab:from-scratch}
    \begin{center}
    \small
        \begin{tabular}{lcc}
            \toprule
            \textbf{Environment} & \textbf{Supervised (1K trans)} & \textbf{DLHF (1M prefs)} \\
            \midrule
            Maze $10 \times 10$
            & $0.0003 \pm 0.0001$
            & $\mathbf{0.0001 \pm 0.0000}$ \\
            Sliding $5 \times 5$
            & $0.6622 \pm 0.0347$
            & $\mathbf{0.0105 \pm 0.0023}$ \\
            Sokoban
            & $0.0198 \pm 0.0018$
            & $\mathbf{0.0181 \pm 0.0016}$ \\
            2048
            & $1.2063 \pm 0.0121$
            & $\mathbf{1.2024 \pm 0.0452}$ \\
            \bottomrule
        \end{tabular}
    \end{center}
\end{table}


\subsection{Sample Efficiency of Preference-Based Dynamics Learning}
\label{sec:sample-efficiency}

We have seen that DLHF can learn simple world models under idealized settings. Unfortunately, even with a noiseless oracle providing preferences, an outsize number of preferences are needed to match supervised performance. This limitation means that naive DLHF is unsuited for most practical learning problems. A natural next question is thus whether DLHF has a more sample efficient instantiation. In this section, we show that RENEW, which directs preference queries toward high-uncertainty transitions via ensemble disagreement, is one such algorithm.

We compare naive DLHF and RENEW on Sliding Tile ($3 \times 3$ and $5 \times 5$), Sokoban, and Maze ($5 \times 5$ and $10 \times 10$). Both methods start from the same pretrained checkpoint and use identical architecture, optimizer, and ensemble settings (\cref{app:architecture}). The only difference is start-state selection: naive samples uniformly, RENEW selects proportional to ensemble disagreement. We use $K = 2$ candidates and batch size $B = 64$, giving both methods 64 preference labels per gradient step. Results report mean $\pm$ 95\% confidence intervals over 5 seeds.

\begin{figure}[t]
    \centering
    \includegraphics{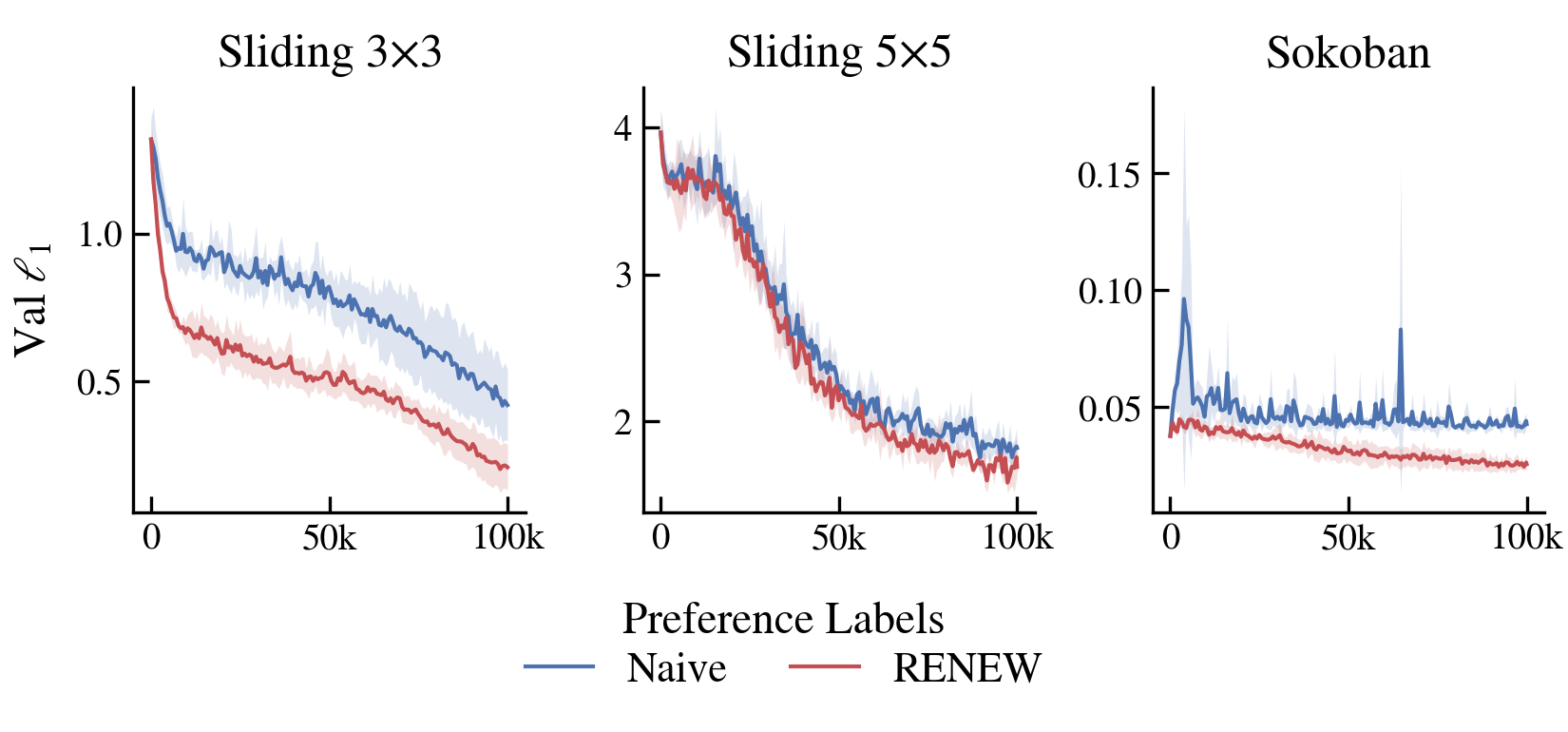}
    \caption{Validation $\ell_1$ error versus cumulative preference labels for naive DLHF and RENEW across three Jumanji environments. Mean over 5 seeds; shaded regions indicate 95\% CIs.}
    \label{fig:sample-efficiency}
\end{figure}

\begin{table}[h]
    \caption{Final validation $\ell_1$ after 100K preference labels. Mean $\pm$ 95\% CI over 5 seeds.}
    \label{tab:sample-efficiency}
    \begin{center}
    \footnotesize
    \setlength{\tabcolsep}{3pt}
        \begin{tabular}{l ccccc}
            & \multicolumn{2}{c}{\bf Sliding Tile} & \multirow{2}{*}{\bf Sokoban} & \multicolumn{2}{c}{\bf Maze} \\
            \cline{2-3} \cline{5-6} \\
            \bf Method & $\mathbf{3 \times 3}$ & $\mathbf{5 \times 5}$ & & $\mathbf{5 \times 5}$ & $\mathbf{10 \times 10}$
            \\ \hline \\
            Pretrained & $1.3200 \pm 0.0510$ & $3.9660 \pm 0.1030$ & $0.0377 \pm 0.0012$ & $0.0233 \pm 0.0033$ & $0.0060 \pm 0.0005$ \\
            Naive      & $0.4190 \pm 0.1220$ & $1.8190 \pm 0.0370$ & $0.0428 \pm 0.0016$ & $0.0019 \pm 0.0006$ & $0.0001 \pm 0.0000$ \\
            RENEW      & $\mathbf{0.2070 \pm 0.0790}$ & $\mathbf{1.6890 \pm 0.1070}$ & $\mathbf{0.0257 \pm 0.0025}$ & $\mathbf{0.0018 \pm 0.0006}$ & $\mathbf{0.0001 \pm 0.0000}$ \\
        \end{tabular}
    \end{center}
\end{table}

\Cref{fig:sample-efficiency} and \cref{tab:sample-efficiency} show that RENEW's active querying substantially improves sample efficiency over naive DLHF across discrete environments, roughly halving the final error on Sliding Tile $3 \times 3$ under the same preference budget. This is consistent with the well-established finding that active query selection improves sample efficiency in preference learning \citep{christiano2017deep}. However, one unexpected result also emerged: on Sokoban, naive DLHF degrades the pretrained model while RENEW improves it, suggesting that active querying may also reduce catastrophic forgetting of pretrained dynamics. We expand on this observation in the next subsection.

\subsection{Repairing Model Exploitation}
\label{sec:repair}

Our final question is whether RENEW can repair model exploitation. Exploitation is essentially unavoidable for imperfect models \citep{bhamidipaty2026imperfect}, but in practice it is concentrated in regions of high epistemic uncertainty \citep{yu2020mopo, kidambi2020morel}. We measure whether RENEW reduces both prediction error and epistemic uncertainty in pretrained world models more effectively than naive finetuning.

\begin{figure}[t]
    \centering
    \includegraphics[]{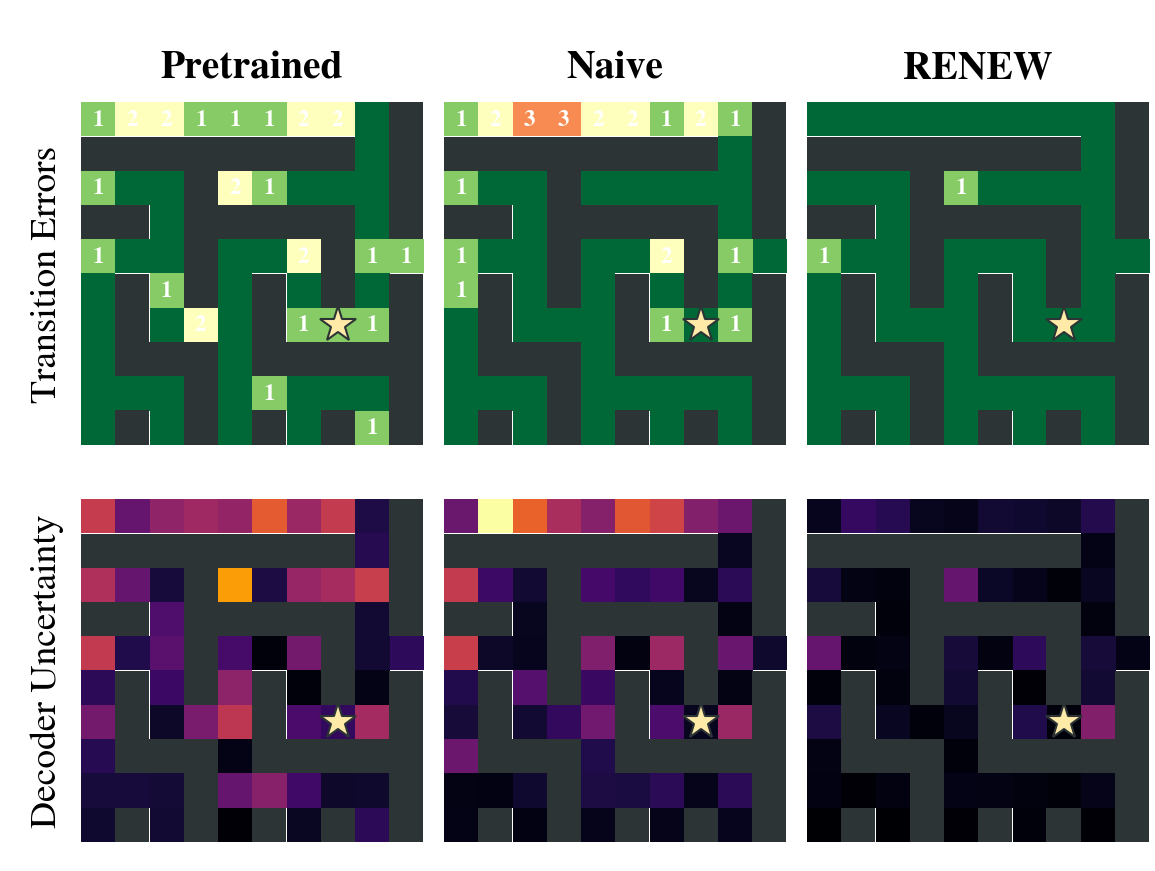}
    \caption{A single Maze $10\times10$ instance (matched 1600-label budget). \textbf{Top:} per-cell transition error (green cells are correct, warmer cells are mispredictions labeled by the number of wrong transitions). \textbf{Bottom:} per-cell epistemic uncertainty (darker = lower/better). Both finetuned models improve over the pretrained baseline (pretrained: $86.8\%$, naive: $88.6\%$, RENEW: $99.1\%$), though naive forgets a few transitions the pretrained model had already predicted correctly.}
    \label{fig:triptych}
\end{figure}

We pretrain on Maze environments of sizes $5 \times 5$, $10 \times 10$, $15 \times 15$, and $20 \times 20$, with each ensemble member trained on 500 offline transitions. Both methods finetune on 1600 preference labels using the same setup as \cref{sec:sample-efficiency}, except with $K = 4$ candidates and 3 active rounds.\footnote{An ablation over $K$ (\cref{tab:ablation_k}) shows $K = 2$ is optimal under a fixed label budget, suggesting the results reported here may understate RENEW's advantage.}

\Cref{fig:triptych} visualizes the effect on a single Maze $10 \times 10$ instance. The pretrained model (left) exhibits high epistemic uncertainty in regions where it makes prediction errors. Naive finetuning (center) reduces some errors but leaves substantial uncertainty. RENEW (right) reduces both more effectively by concentrating supervision where the model is most wrong. \Cref{tab:maze_accuracy} confirms this quantitatively across maze sizes: RENEW outperforms naive finetuning at every size under the same preference budget, with tighter confidence intervals. Both methods substantially improve over the pretrained baseline, confirming that preference labels can repair model errors without additional demonstrations.

\begin{table}[h]
    \caption{Transition accuracy (\%) across Maze sizes. Preference budget: 1600 labels. Mean $\pm$ 95\% CI over 10 seeds.}
    \label{tab:maze_accuracy}
    \begin{center}
        \begin{tabular}{lcccc}
            \multicolumn{1}{l}{\bf Method} & \multicolumn{1}{c}{\bf 5$\boldsymbol{\times}$5} & \multicolumn{1}{c}{\bf 10$\boldsymbol{\times}$10} & \multicolumn{1}{c}{\bf 15$\boldsymbol{\times}$15} & \multicolumn{1}{c}{\bf 20$\boldsymbol{\times}$20}
            \\ \hline \\
            Pretrained       & $83.8 \pm 3.6$ & $87.6 \pm 2.1$ & $87.8 \pm 2.8$ & $87.9 \pm 2.0$ \\
            Naive            & $89.7 \pm 3.5$ & $93.2 \pm 2.4$ & $92.3 \pm 1.7$ & $93.2 \pm 2.3$ \\
            RENEW            & $\mathbf{96.5 \pm 1.5}$ & $\mathbf{97.2 \pm 1.4}$ & $\mathbf{94.4 \pm 1.6}$ & $\mathbf{94.9 \pm 1.3}$ \\
        \end{tabular}
    \end{center}
\end{table}

This observation also suggests a mechanism for the catastrophic forgetting observed in \cref{sec:sample-efficiency}: uniform sampling risks overwriting correct pretrained dynamics with updates from inaccurate imagined transitions, while RENEW avoids this by concentrating supervision on high-uncertainty regions where the model is most likely wrong.

\section{Conclusion}

This paper introduced Dynamics Learning from Human Feedback (DLHF), a framework for learning world model dynamics directly from binary preferences, and RENEW, an algorithm that uses epistemic uncertainty to direct preference queries toward transitions where the model is most vulnerable to exploitation. We showed that binary preferences can serve as a training signal for latent dynamics world models, though at substantially higher label cost than supervised learning; that RENEW improves sample efficiency over naive DLHF and avoids catastrophic forgetting of pretrained dynamics; and that preference-based finetuning reduces prediction error and epistemic uncertainty in regions vulnerable to model exploitation.

The core novelty is a shift in what preferences supervise. Prior work uses preferences to learn reward functions or align policies; DLHF uses them to learn transition dynamics directly, requiring no reward labels, no expert demonstrations, and no environment interaction.

Several limitations remain. Our experiments use a synthetic oracle rather than real human annotators, and the environments we evaluate are small discrete grid worlds and low-dimensional continuous control tasks. Validating DLHF with noisy human labels and scaling to higher-dimensional observation spaces are the most important next steps. Connecting empirical reductions in prediction error to formal guarantees of unexploitability \citep{bhamidipaty2026imperfect} and practical offline RL algorithms \citep{jackson2026clean} are other important directions.


\bibliography{rlj}
\bibliographystyle{rlj}

\beginSupplementaryMaterials
\appendix
\crefalias{section}{appendix}
\crefalias{subsection}{appendix}
\crefalias{subsubsection}{appendix}

\section{Preliminaries}
\label{sec:prelims}

We briefly review the necessary background to understand DLHF and RENEW.

\subsection{Reinforcement Learning}
We consider a Markov decision process (MDP) $\mathcal{M} = (\mathcal{S}, \mathcal{A}, \mathcal{T}, d_0, R, \gamma)$, where $\mathcal{S}$ is the state space, $\mathcal{A}$ is the action space, $\mathcal{T}: \mathcal{S} \times \mathcal{A} \to \Delta(\mathcal{S})$ is the dynamics model, $d_0$ is the initial state distribution, $R: \mathcal{S} \times \mathcal{A} \to \mathbb{R}$ is the reward function, and $\gamma \in [0,1)$ is the discount factor. A policy $\pi(a|s)$ induces a distribution over trajectories. We use $\tau$ to denote trajectories and $\sigma$ to denote trajectory segments.

\subsection{Model-Based RL, World Models, and Model Exploitation}
\label{sec:mbrl}

In model-based RL (MBRL), the agent learns an approximate dynamics model $\hat{\mathcal{T}}_\theta(s' \mid s, a)$ from observed transitions, typically via maximum likelihood:
\begin{equation}
\label{eq:mle}
    \theta^* \in \arg\max_\theta \, \mathbb{E}_{(s,a,s') \sim \mathcal{D}}\left[\log \hat{\mathcal{T}}_\theta(s' \mid s, a)\right].
\end{equation}
The agent then plans within this learned model, generating synthetic rollouts to optimize its policy without additional environment interaction \citep{dyna}. This is what makes MBRL substantially more sample efficient than model-free methods, which must update value functions or policies directly from real transitions \citep{deisenroth2011pilco, chua2018deep, janner2019trust}.

We refer to $\hat{\mathcal{T}}_\theta$ broadly as a \textit{world model} \citep{ha2018world}. World models may operate directly in the state space or over learned latent representations $z$, where an encoder $e_\psi(z \mid s)$ maps observations to a latent space and a latent dynamics model $\hat{\mathcal{T}}_\theta(z' \mid z, a)$ predicts transitions in that space \citep{hafner2019learning}. World models may also predict rewards, but this work focuses exclusively on dynamics.

The central vulnerability of MBRL is \textit{model exploitation}: because the learned dynamics model is imperfect, a policy optimizing within it can discover transitions that $\hat{\mathcal{T}}_\theta$ misestimates and exploit them to achieve artificially high returns that do not transfer to the true environment \citep{ha2018world, janner2019trust, bhamidipaty2026imperfect}.

\subsection{Offline Reinforcement Learning}
\label{sec:offline}

In offline RL, the agent learns entirely from a fixed dataset $\mathcal{D}$ of transitions collected by one or more behavior policies, with no further environment interaction \citep{tutorial}. This creates a distributional shift problem: the policy being learned may need to act in states or take actions that $\mathcal{D}$ covers poorly, and without the ability to collect corrective data, errors in these regions compound \citep{kumar2019stabilizing, fujimoto2019off}. For model-based offline methods, distributional shift manifests directly as model exploitation: the world model is trained on $\mathcal{D}$ and is unreliable outside its support, so a policy planning in the model will find and exploit exactly those unreliable transitions.

We consider the reward-free offline setting, where $\mathcal{D}$ contains only transition tuples $\{(s_i, a_i, s_i')\}_{i=1}^N$ without reward labels. This is practically motivated: in domains such as robotics, autonomous driving, and healthcare, large corpora of interaction data exist but reward annotation is expensive or unavailable \citep{zolna2020offline, yu2022leverage}.

\subsection{Reinforcement Learning from Human Feedback}
\label{sec:rlhf}

Reinforcement learning from human feedback (RLHF) is a class of methods for learning reward functions from human judgments, typically in the form of pairwise preferences over trajectory segments \citep{christiano2017deep}. The objective is to optimize a reward model $R_\phi$ under the Bradley-Terry preference model \citep{bradley1952rank} given a dataset of pairwise preferences $\mathcal{D}_\succ = \{(\sigma^0, \sigma^1, y)\}$ where $(\sigma^0, \sigma^1)$ is a trajectory segment pair and $y \in \{0, 1, 0.5\}$ is a preference label indicating whether $\sigma^0$ is preferred, $\sigma^1$ is preferred, or both are equally preferred, respectively. The Bradley-Terry model estimates the probability that $\sigma^0$ is preferred over $\sigma^1$ as:
\begin{equation}
\label{eq:bt}
    P(\sigma^0 \succ \sigma^1) = \frac{\exp R_\phi(\sigma^0)}{\exp R_\phi(\sigma^0) + \exp R_\phi(\sigma^1)} = \mathrm{logistic}\!\left(R_\phi(\sigma^0) - R_\phi(\sigma^1)\right)
\end{equation}
where $\mathrm{logistic}(x) = 1 / ({1 + \exp(-x)})$. The reward model parameters $\phi$ are optimized by minimizing the cross-entropy loss over the preference dataset:
\begin{equation}
\label{eq:rlhf_loss}
    \mathcal{L}(\phi) = -\mathbb{E}_{(\sigma^0, \sigma^1, y)}\left[(1 - y) \log P(\sigma^0 \succ \sigma^1) + y \log P(\sigma^1 \succ \sigma^0)\right].
\end{equation}
While RLHF has typically been applied to learn reward functions or policies \citep{christiano2017deep, ouyang2022training}, in this work we apply the preference learning framework to the \textit{dynamics model} $\hat{\mathcal{T}}_\theta$, using human feedback to improve the fidelity of learned world models rather than to specify task objectives.

\section{Related Work}

\paragraph{Learning dynamics from preferences.} Several recent papers have detailed methods for improving world model realism with human preferences. \citet{cao2024reinforcement} use preferences from licensed drivers to train a reward model that aligns traffic simulation agents with human judgments of realism, and \citet{katz2019learning} use pilot preferences to tune an airspace encounter model via inverse reinforcement learning. In both cases, the preference signal supervises a reward model rather than the dynamics model itself, and the approach is tailored to a specific domain with access to expert labelers. \citet{andrus} formalizes the inverse problem of learning dynamics from demonstrations but similarly assumes access to the reward function. In contrast, DLHF is, to our knowledge, the first formulation for learning dynamics directly from preferences, requiring no reward labels, no demonstrations, and no environment interaction.

\paragraph{Offline model-based RL.} Model exploitation in offline model-based RL has primarily been addressed through pessimism. MOPO \citep{yu2020mopo} penalizes predicted returns by ensemble disagreement, discouraging the policy from visiting states where the model is uncertain. MOReL \citep{kidambi2020morel} halts rollouts when uncertainty exceeds a threshold, restricting planning to well-covered regions. COMBO \citep{yu2021combo} regularizes value estimates on out-of-distribution state-action pairs without explicit uncertainty quantification. All three approaches trade coverage for robustness, limiting the ability of model-based methods to generalize beyond the training distribution. DAgger \citep{dagger} takes a different approach, collecting corrective expert labels in regions where the learned model fails, but requires environment interaction and access to an expert policy. Sim-OPRL \citep{pace2024preference} is closest to our setup: it operates offline, learns a transition model, generates rollouts in simulation, and elicits preferences, but uses those preferences exclusively to learn a reward model while leaving the dynamics fixed. RENEW offers a third response to model exploitation: rather than avoiding uncertain regions or collecting additional demonstrations, it repairs them directly using preference supervision directed by epistemic uncertainty.

\paragraph{World models and physical plausibility.} A growing body of work has examined whether learned world models capture physically plausible dynamics. \citet{kang2024far} show that video generation models trained at scale fail to extract general physical rules, instead exhibiting case-based generalization. \citet{motamed2026generative} confirm that even large-scale generative world models violate solidity, object persistence, and basic mechanics, and WorldModelBench \citep{li2025worldmodelbench} demonstrates through 67,000 human annotations that these violations are both pervasive and reliably judged by humans. These findings motivate preference-based supervision: the gap between what MLE learns from limited data and what humans trivially recognize as wrong is precisely the gap RENEW targets. RLVR-World \citep{wu2025rlvr} also identifies MLE as a misaligned training objective but addresses it with verifiable task-specific rewards that require ground-truth access. RENEW instead uses binary preferences to close dynamics fidelity gaps without ground-truth states, targeting the complementary problem of physical plausibility rather than task alignment.

\section{Architecture and Optimization Details}
\label{app:architecture}

We use two latent dynamics world model architectures: a convolutional model that outputs categorical distributions over discrete tile states for Jumanji environments (\cref{app:conv-arch}), and a Gaussian MLP that outputs diagonal Gaussian distributions over continuous next states for classic control environments (\cref{app:gaussian-mlp-arch}). Both define a trajectory log-likelihood that serves as the preference function in the DLHF loss (\cref{eq:dlhf_loss}). All experiments were run on a single NVIDIA GeForce RTX 2080 Ti GPU. \Cref{tab:hyperparams} reports hyperparameters.

\begin{table}[h]
    \caption{Hyperparameters for both architectures.}
    \label{tab:hyperparams}
    \begin{center}
    \small
        \begin{tabular}{lcc}
            & \bf Jumanji & \bf Classic Control \\
            \hline \\
            Ensemble size $E$       & 3             & 3 \\
            Batch size $B$          & 64            & 64 \\
            Segment horizon $H$     & 1             & 1 \\
            BT temperature $\beta$  & 1.0           & 1.0 \\
            Candidate pool size     & 1{,}024       & 1{,}024 \\
            Optimizer               & Adam          & Adam \\
            Learning rate           & $3 \times 10^{-4}$ & $3 \times 10^{-3}$ \\
        \end{tabular}
    \end{center}
\end{table}

\subsection{Convolutional Latent Dynamics Model}
\label{app:conv-arch}

The world model consists of three components: an encoder, a latent dynamics model, and a decoder, all operating on a 2D grid representation. We use ConvNeXt blocks~\citep{liu2022convnet} as the residual unit throughout. Each block applies a depthwise $3 \times 3$ convolution, LayerNorm, a two-layer MLP with $4\times$ channel expansion and GELU activation, and a residual connection.

The encoder maps discrete grid observations to a latent representation. Each cell is embedded via a learned tile embedding of dimension 32, producing a $(\text{rows}, \text{cols}, 32)$ grid. A linear projection expands this to 64 latent channels, followed by 3 ConvNeXt blocks. The dynamics model predicts the next latent state given the current latent and an action. The action is one-hot encoded and projected to a $(\text{rows}, \text{cols}, 64)$ map via a linear layer, concatenated channel-wise with the latent, projected back to 64 channels, and processed through 4 ConvNeXt blocks. The decoder predicts per-cell tile logits from the next latent state and action using the same action injection, 3 ConvNeXt blocks, and a final linear layer outputting $(\text{rows}, \text{cols}, N_{\text{tiles}})$ logits. During pretraining, the model is trained with cross-entropy over tile classes. During preference finetuning, the log-softmax of these logits provides the trajectory log-likelihood.

\subsection{Gaussian MLP}
\label{app:gaussian-mlp-arch}

The world model concatenates the state vector with a one-hot encoding of the discrete action and passes the result through two fully connected layers of 128 units with tanh activations. The output head predicts the mean of the next state. A separate learned parameter vector stores the log-standard-deviation per action. The model defines a diagonal Gaussian over next states, and the trajectory log-likelihood is the sum of per-dimension Gaussian log-probabilities.

\section{Practical Considerations}
\label{sec:practical-considerations}

Following \citet{christiano2017deep}, we highlight several implementation details that we found important for RENEW to work well in practice.

\paragraph{Single-transition segments.} We set the segment horizon to $H = 1$, reducing each preference query to a single-transition comparison. This avoids cascading prediction errors over long rollouts that can make trajectory-level preferences noisy and uninformative. At $H = 1$, each query asks the simplest possible question: \textit{which single transition is more plausible?} In the discrete environments we evaluate, single transitions produce perceptible changes that are easy to judge. In environments with smoother dynamics or long-horizon dependencies, longer segments may be necessary to surface distinguishable differences between model and oracle.

\paragraph{Relaxed action sequences.} While the formulation in \cref{alg:renew} is agnostic to how segment pairs are constructed, a natural choice is to roll out the same start state and action sequence through both the world model and an oracle, forming a pair from the two resulting segments. In practice, we find that action sequences need not match across pairs: the Bradley-Terry loss compares log-likelihoods of fixed segments regardless of how they were generated. This contrasts with typical RLHF, where segment pairs share a common prompt or initial condition to control for context. In DLHF, the preference label reflects absolute plausibility under the true dynamics rather than relative quality conditioned on a shared input, so this constraint can be relaxed.

\paragraph{Ranking-based pair construction.} For each high-uncertainty state-action pair, we sample $K$ candidate transitions, identify the most plausible one, and form $K - 1$ binary preference pairs against the remaining candidates. This follows the ranking-based elicitation strategy used in InstructGPT \citep{ouyang2022training}, which extracts more training signal per query by converting a single $K$-way ranking into multiple pairwise comparisons.

\paragraph{Differentiable sampling.} Note that while generating trajectory segments from $\hat{\mathcal{T}}_\theta$ involves non-differentiable sampling, $\mathcal{L}_{\text{DLHF}}$ is computed over fixed trajectory segments and is fully differentiable with respect to $\theta$. The sampling step could be made differentiable via the reparameterization trick \citep{kingma2013auto} for continuous dynamics or Gumbel-Softmax relaxation \citep{jang2017categorical, maddison2017concrete} for discrete dynamics, enabling end-to-end gradient flow through rollout generation. In practice, we found this unnecessary: optimizing the log-likelihood of fixed segments provides a sufficient training signal, and differentiable sampling did not improve convergence.

\section{Jumanji Environments}
\label{app:envs}

For reference, we describe some of the Jumanji \citep{bonnet2024jumanji} environments used in our experimental evaluation.

\begin{itemize}
    \item \textbf{Maze $10 \times 10$.} A grid world where an agent navigates corridors to reach a goal position. Walls are procedurally generated; the agent can move in four cardinal directions.
    \item \textbf{Sliding Tile $5 \times 5$.} A generalization of the classic 15-puzzle to a $5 \times 5$ board. Tiles are shuffled from the solved configuration, and the player slides tiles into the single empty cell to restore the original ordering.
    \item \textbf{Sokoban.} A puzzle game where the agent pushes boxes onto designated target cells. Boxes can only be pushed, not pulled, and cannot pass through walls or other boxes, making many moves irreversible.
    \item \textbf{2048.} A single-player tile-merging game on a $4 \times 4$ grid. The player slides all tiles in a cardinal direction; matching tiles merge and double in value. After each valid move, a new tile spawns in a random empty cell. Unlike the other environments, 2048 has stochastic dynamics.
\end{itemize}

\section{Ablation Studies}

We ablate two design choices in RENEW: the number of candidate rollouts $K$ used for ranking-based pair construction, and the treatment of tied preferences in the Bradley-Terry loss. All ablations use a fixed budget of 10K preference labels, an ensemble of $E = 3$ members, and report mean $\pm$ 95\% CI over 5 seeds.

\subsection{Number of Candidates ($K$)}

For each start state, RENEW samples $K$ candidate transitions from the model and forms $K - 1$ preference pairs by comparing each candidate against the best (\cref{sec:practical-considerations}). Larger $K$ extracts more training signal per oracle query but reduces the number of training steps under a fixed label budget, since each step consumes $B(K-1)$ labels. We sweep $K$ over $\{2, 3, 4, 6, 8\}$ on maze10, sliding3, and sokoban. For each value of $K$, the naive baseline uses a batch size of $B(K-1)$ to match RENEW's labels per step, isolating active selection as the only variable.


\begin{table}[h]
    \caption{Effect of $K$ on final validation $\ell_1$. Each naive baseline uses batch size $B(K-1)$ to match RENEW's labels per step. Budget: 10K labels, $E = 3$. Mean $\pm$ 95\% CI over 5 seeds.}
    \label{tab:ablation_k}
    \begin{center}
    \small
        \begin{tabular}{llccccc}
            & & \multicolumn{5}{c}{\bf $K$} \\
            \cline{3-7} \\
            \bf Env & \bf Method & \bf 2 & \bf 3 & \bf 4 & \bf 6 & \bf 8
            \\ \hline \\
            \multirow{2}{*}{maze10}
            & Naive  & $\mathbf{.022 \pm .004}$ & $.026 \pm .004$ & $.030 \pm .003$ & $.032 \pm .002$ & $.035 \pm .003$ \\
            & RENEW  & $\mathbf{.021 \pm .001}$ & $.026 \pm .003$ & $.030 \pm .003$ & $.034 \pm .003$ & $.041 \pm .008$ \\
            \hline \\
            \multirow{2}{*}{sliding3}
            & Naive  & $\mathbf{1.26 \pm .03}$ & $1.48 \pm .04$ & $1.58 \pm .08$ & $1.81 \pm .10$ & $1.94 \pm .10$ \\
            & RENEW  & $\mathbf{1.29 \pm .09}$ & $1.50 \pm .08$ & $1.65 \pm .08$ & $1.87 \pm .08$ & $2.03 \pm .05$ \\
            \hline \\
            \multirow{2}{*}{sokoban}
            & Naive  & $\mathbf{.090 \pm .008}$ & $.124 \pm .017$ & $.140 \pm .011$ & $.186 \pm .048$ & $.215 \pm .069$ \\
            & RENEW  & $\mathbf{.079 \pm .016}$ & $.124 \pm .027$ & $.129 \pm .010$ & $.174 \pm .014$ & $.222 \pm .042$ \\
        \end{tabular}
    \end{center}
\end{table}

\Cref{tab:ablation_k} shows that $K = 2$ achieves the lowest $\ell_1$ error for both methods across all three environments, with performance degrading monotonically as $K$ increases. Under a fixed budget of 10K labels, the gradient step reduction at large $K$ dominates the benefit of richer per-query signal: at $K = 2$ the model receives 156 gradient updates, while at $K = 8$ it receives 22.

The gap between RENEW and the naive baseline is small at every value of $K$, contrasting with the finetuning results in \cref{sec:repair} where RENEW shows a clear advantage. This may reflect a difference in the structure of epistemic uncertainty when finetuning versus training from scratch. When finetuning, uncertainty is concentrated in regions where the offline data provided poor coverage, giving RENEW clear targets for active selection. When training from scratch, uncertainty is approximately uniform everywhere, so active selection offers little advantage over uniform sampling. We leave verification of this hypothesis to future work.

\subsection{Tie Exclusion}

When two candidate rollouts have identical $\ell_1$ error to the ground truth, the oracle has no preference. By default, RENEW excludes these tied pairs from the Bradley-Terry loss. The alternative is to include them with a preference label of $0.5$, which pushes the model toward assigning equal likelihood to both candidates. We compare both strategies for RENEW and the batch-matched naive baseline on \texttt{maze10}, \texttt{sliding3}, and \texttt{sokoban} with $K = 2$.

\begin{table}[h]
    \caption{Effect of tie exclusion on final validation $\ell_1$. Budget: 10K labels, $K = 2$, $E = 3$. Mean $\pm$ 95\% CI over 5 seeds.}
    \label{tab:ablation_ties}
    \begin{center}
    \small
        \begin{tabular}{llcc}
            \bf Env & \bf Method & \bf Exclude ties & \bf Include ties
            \\ \hline \\
            \multirow{2}{*}{\texttt{maze10}}
            & Naive  & $.023 \pm .002$ & $\mathbf{.021 \pm .001}$ \\
            & RENEW  & $.022 \pm .002$ & $\mathbf{.022 \pm .001}$ \\
            \hline \\
            \multirow{2}{*}{\texttt{sliding3}}
            & Naive  & $\mathbf{1.26 \pm .03}$ & $1.27 \pm .02$ \\
            & RENEW  & $1.32 \pm .07$ & $\mathbf{1.27 \pm .07}$ \\
            \hline \\
            \multirow{2}{*}{\texttt{sokoban}}
            & Naive  & $.115 \pm .025$ & $\mathbf{.086 \pm .018}$ \\
            & RENEW  & $\mathbf{.078 \pm .020}$ & $.085 \pm .025$ \\
        \end{tabular}
    \end{center}
\end{table}

\Cref{tab:ablation_ties} shows that the effect of tie handling is small and inconsistent across environments and methods. Including ties helps naive on \texttt{maze10} and \texttt{sokoban} but is neutral on \texttt{sliding3}. For RENEW, excluding ties helps on \texttt{sokoban} but including ties helps on \texttt{sliding3}. No consistent pattern emerges, and most differences fall within confidence intervals. We retain tie exclusion as the default for RENEW since it avoids injecting uninformative gradient signal, but the results suggest the method is not sensitive to this choice.


\section{Continuous Control Experiments}
\label{app:classic-control}

To evaluate whether DLHF extends beyond discrete environments, we apply the same comparisons from the main text to MountainCar-v0 (2-dimensional state, 3 actions) and Acrobot-v1 (6-dimensional state, 3 actions) from Gymnax~\citep{gymnax2022github}. Both have continuous state spaces and deterministic dynamics. The world model is a Gaussian MLP predicting diagonal Gaussian distributions over next states (\cref{app:gaussian-mlp-arch}).

Learning continuous dynamics from preferences is inherently harder than the discrete case. In discrete environments, two candidate predictions often land on entirely different tiles, producing an unambiguous preference signal. In continuous environments, both candidates are sampled from the same predicted Gaussian and tend to cluster near the mean, yielding more near-tied comparisons and less informative gradient updates. Despite this, DLHF learns continuous dynamics at near-parity with supervised baselines on both tasks.

\subsection{Learning from Scratch}

We train both supervised and DLHF models from scratch using the same protocol as \cref{sec:from-scratch}. On MountainCar, both methods achieve near-zero MSE: supervised attains $3.1 \times 10^{-5} \pm 8.0 \times 10^{-6}$ and DLHF attains $1.3 \times 10^{-4} \pm 3.7 \times 10^{-5}$ (mean $\pm$ 95\% CI over 6 seeds). Both are effectively perfect predictions of the deterministic dynamics, with supervised slightly lower in absolute terms. On Acrobot, DLHF matches supervised performance within confidence intervals ($15.66 \pm 0.93$ versus $15.63 \pm 0.75$), confirming that preferences can train continuous dynamics models on a 6-dimensional state space.

\subsection{Sample Efficiency}

We compare naive DLHF and RENEW on both environments using an ensemble of $E = 3$ members with a budget of 100K preference labels over 10 seeds. Both methods are warm-started from a shared initialization trained on 5 random transitions for 50 gradient steps. We report results at both $K = 2$ and $K = 8$.

\begin{figure}[h]
    \centering
    \includegraphics[width=0.9\textwidth]{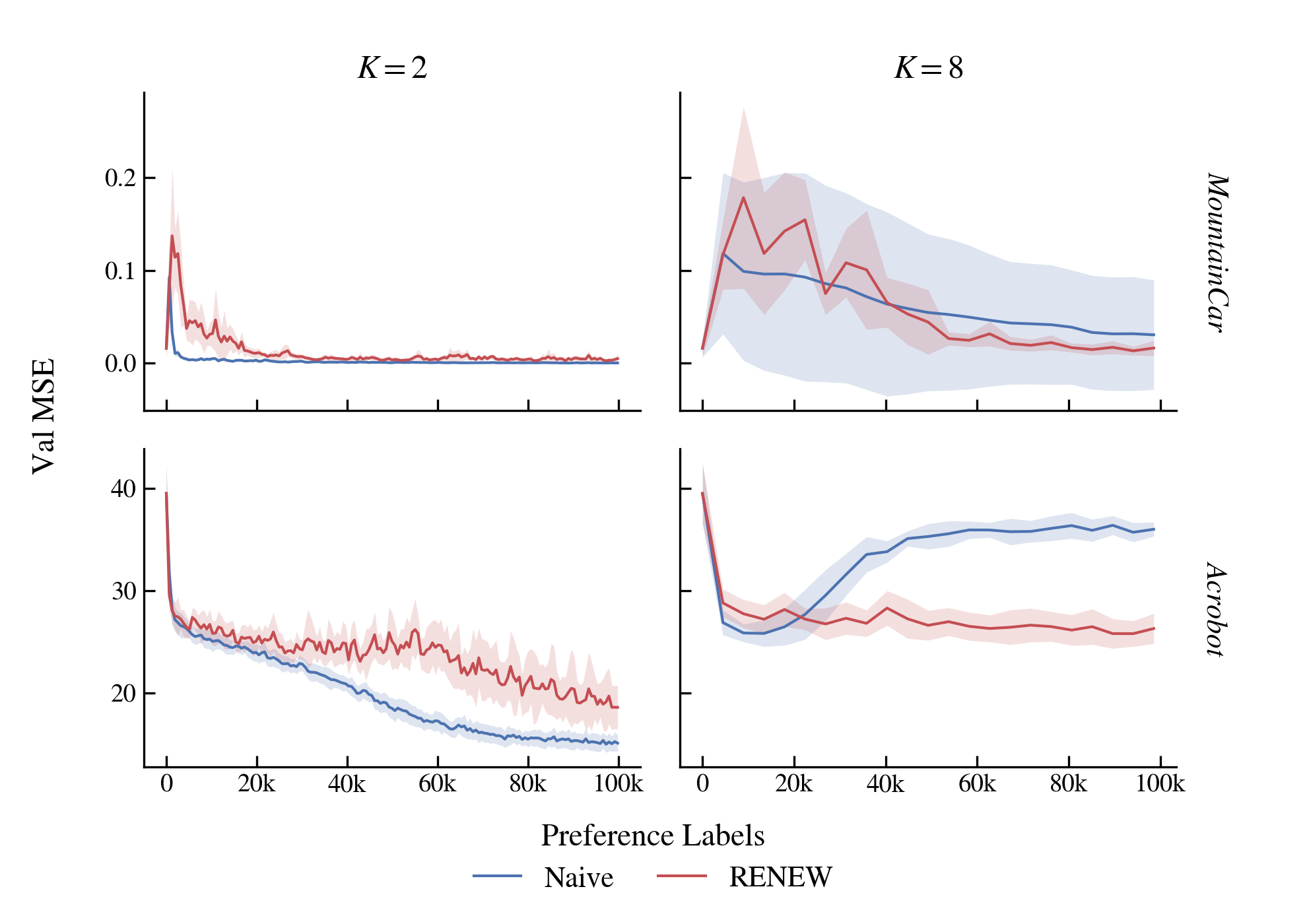}
    \caption{Naive DLHF versus RENEW on MountainCar-v0 and Acrobot-v1 at $K = 2$ (left) and $K = 8$ (right). Solid lines show the mean over 10 seeds; shaded regions indicate 95\% confidence intervals. Both methods improve over the warm-start initialization, but neither attains the low validation MSE seen in the discrete Jumanji environments, reflecting the weaker preference signal from Gaussian versus categorical log-likelihoods.}
    \label{fig:continuous}
\end{figure}

\begin{table}[h]
    \caption{Final validation MSE on classic control
    environments. Both methods are warm-started from the same
    initialization (50 steps on 5 random transitions). Budget:
    100K labels. Mean $\pm$ 95\% CI over 10 seeds. Best per
    setting in \textbf{bold}.}
    \label{tab:classic-control-full}
    \begin{center}
    \small
        \begin{tabular}{ll ccc}
            & & \bf Warm-start & \bf Naive & \bf RENEW \\
            \hline \\
            \multirow{2}{*}{MountainCar-v0}
            & $K = 2$ & \multirow{2}{*}{$.0162 \pm .0066$}
              & $\mathbf{.0004 \pm .0002}$
              & $.0050 \pm .0047$ \\
            & $K = 8$ &
              & $.0306 \pm .0589$
              & $\mathbf{.0164 \pm .0082}$ \\
            \hline \\
            \multirow{2}{*}{Acrobot-v1}
            & $K = 2$ & \multirow{2}{*}{$39.48 \pm 2.03$}
              & $\mathbf{15.09 \pm 0.81}$
              & $18.59 \pm 2.13$ \\
            & $K = 8$ &
              & $35.98 \pm 0.69$
              & $\mathbf{26.28 \pm 1.48}$ \\
        \end{tabular}
    \end{center}
\end{table}

At $K = 2$ (1{,}562 gradient steps), naive outperforms RENEW on both environments. The state-action spaces are small (2 or 6 state dimensions, 3 actions), so uniform sampling provides adequate coverage, and ensemble disagreement is a noisier estimator of model error in continuous spaces than in discrete ones, where member agreement reliably indicates correctness. At $K = 8$ (223 gradient steps), RENEW substantially outperforms naive on Acrobot ($26.28$ vs $35.98$) and preserves the warm-start performance on MountainCar ($0.016$ vs $0.031$ for naive). With fewer gradient steps, each update must count, and RENEW's targeted selection avoids wasting scarce steps on uninformative examples. This is consistent with the catastrophic forgetting pattern observed in the Jumanji experiments: naive degrades the warm-start dynamics on both environments at $K = 8$, while RENEW avoids this.

\end{document}